\documentclass[runningheads]{llncs}

\pdfoutput=1
\usepackage{graphicx}
\usepackage{hyperref}
\hypersetup{
    colorlinks=true,
    citecolor=black,
    linkcolor=black,
    filecolor=black,
    urlcolor=blue,
}
\usepackage[inline]{enumitem}

\usepackage[super]{nth}
\usepackage{algpseudocode}
\usepackage{makecell}
\usepackage[cmex10]{amsmath}
\usepackage{amsfonts}
\usepackage{amssymb}

\usepackage{algorithm}
\usepackage{array}

\usepackage{url}
\usepackage{mdwmath}
\usepackage{mdwtab}
\usepackage{eqparbox}
\usepackage{stfloats}
\usepackage{siunitx}
\usepackage{mwe}    % loads »blindtext« and »graphicx«
\usepackage[caption=false,font=footnotesize]{subfig}

\usepackage{bibentry}
\nobibliography*

\newif\ifdraft
\draftfalse
\begin{document}
\title{Sim-Env: Decoupling OpenAI Gym Environments from Simulation Models
\ifdraft \\ DRAFT \fi}
\titlerunning{Sim-Env}
% If the paper title is too long for the running head, you can set
% an abbreviated paper title here
%
 \author{Andreas Schuderer\inst{1,2} \and
 Stefano Bromuri\inst{1} \and Marko van Eekelen\inst{1,3}
% Third Author\inst{3}\orcidID{2222--3333-4444-5555}}
% %
\authorrunning{A. Schuderer et al.}
% % First names are abbreviated in the running head.
% % If there are more than two authors, 'et al.' is used.
% %
 \institute{Open University of the Netherlands,\\ Valkenburgerweg 177, Heerlen, the Netherlands,\\
 \email{andreas.schuderer@ou.nl}, \email{stefano.bromuri@ou.nl}, \email{marko.vaneekelen@ou.nl}\\
  \and APG Algemene Pensioen Groep N.V.,\\ Oude Lindestraat 70, Heerlen, the Netherlands,
  \and Radboud University,\\ Houtlaan 4, Nijmegen, the Netherlands.
  }
}
 
% \email{\{abc,lncs\}@uni-heidelberg.de}}
%
\maketitle  % typeset the header of the contribution

\begin{abstract}
Reinforcement learning (RL) is one of the most active fields of AI research. Despite the interest demonstrated by the research community in reinforcement learning, the development methodology still lags behind, with a severe lack of standard APIs to foster the development of RL applications. OpenAI Gym is probably the most used environment to develop RL applications and simulations, but most of the abstractions proposed in such a framework are still assuming a semi-structured methodology. This is particularly relevant for agent-based models whose purpose is to analyse adaptive behaviour displayed by self-learning agents in the simulation. 
In order to bridge this gap, we present a workflow and tools for the decoupled development and maintenance of multi-purpose agent-based  models and derived single-purpose reinforcement learning environments, enabling the researcher to swap out environments with ones representing different perspectives or different reward models, all while keeping the underlying domain model intact and separate. The \textit{Sim-Env} Python library generates OpenAI-Gym-compatible reinforcement learning environments that use existing or purposely created domain models as their simulation back-ends. Its design emphasizes ease-of-use, modularity and code separation. 

\end{abstract}

\section{Introduction}\label{intro}

% Context

Reinforcement learning is a field of AI concerned with the problem of modelling situated agents which can learn by receiving rewards from an environment \cite{van2016true}, without explicitly programming the behaviour of the agents. The research community has been particularly active in recent years, with applications of RL in traffic control \cite{traffic}, game theory \cite{li2019nonzero}, walking robots \cite{wu2017posture}, swarm optimisation \cite{huttenrauch2019deep}, automatic control \cite{pang2020reinforcement}. This research has even intensified with the advent of deep learning architectures for reinforcement learning \cite{arulkumaran2017deep}. 

In spite of the thriving development from the algorithmic perspective, the software engineering methodology is still lagging behind.

The \textit{OpenAI Gym} initiative has created a platform and programming interface for reinforcement learning (RL) agents to interact with environments \cite{Brockman16}. A significant number of OpenAI-Gym-compatible environments have been made available since, allowing researchers to test and benchmark their RL algorithms.

These existing environments act as accessible and easy-to-use benchmarks, reducing the work needed to carry out reproducible RL research. We argue that, in recent years, this increased availability has helped increase the body of research in the field. For instance, \cite{rl-zoo} present an extensive library of pre-trained agents for OpenAI Gym environments.

% Problem

However, for researchers working on agent-based domain models \cite{niazi2012cognitive} for the purpose of simulations of adaptive behaviours, creating a new Open-AI-Gym-compatible RL environment (or any type of RL environment) for their domain model still requires  custom software development specifically for the purpose of the desired environment. This can lead to code duplication and hard-to-maintain bridging code, particularly in the light of change management and in cases of a non-trivial matrix of research problems and if parameter sweeps in the problem domain are desired.

The main contribution of this work it to propose a workflow which reduces the coupling between the domain model and the RL environment(s), and present the Python library \textit{Sim-Env} as a suitable tool to facilitate this workflow. The researcher is enabled to use an existing or newly developed domain model to generate one or several different RL environments with minimal effort, which may represent different (single- or multi-agent) RL research problems. As the task of creating a thoroughly validated model is a serious investment of time and effort, this decoupling increases the model's re-usability and therefore its value for research.

Additionally, Sim-Env allows researchers to easily create plugins that orthogonally extend domain model capability (for instance, replacing static financial assets with another asset management model). Sharing these plugins with the community adds modular contributions to the public collection of environments that are available to all users of the OpenAI Gym framework.

This approach has several advantages above creating OpenAI-Gym-compatible environments from scratch: While previously, environments usually were inseparable from their domain model, they can now be maintained separately and more granularly. At the same time, this does not detract from creating reproducible results, as the ability to rigorously version specific environments is retained.

In this paper, Section \ref{approach} describes the proposed workflow as well as the design aspects and usage of Sim-Env; Section \ref{eval} demonstrates the viability of Sim-Env on an example; Section \ref{related} positions our work in the field of RL tools; Section \ref{future} summarises our work and discusses relevant future directions.

\section{Background}\label{bg}

\subsection{Reinforcement Learning}

Reinforcement learning (RL) is a machine learning approach to decision making problems. A software agent takes actions which affect the environment. Its inputs at each step in time are some representation of the current state of the environment and the reward signal belonging to the current state of the environment. Its goal is to take a sequence of actions which maximise the cumulative reward received.

\begin{figure}[htb!]
	\centering
	\includegraphics[scale=0.8]{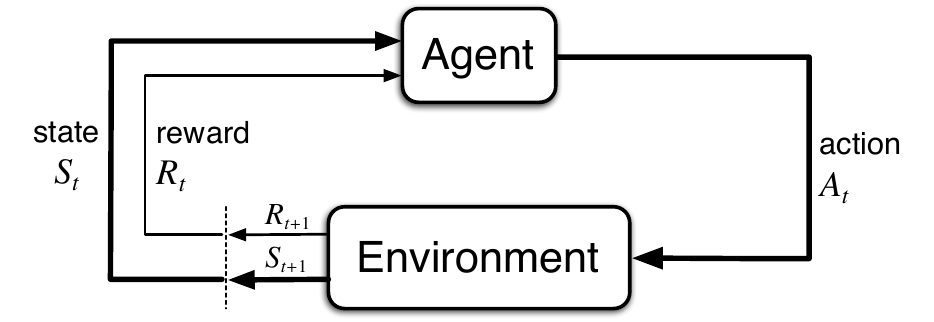}
	\caption{Interaction between the agent and the environment in RL. Reprinted from \textit{Reinforcement learning: An introduction} (p. 48), by R. S. Sutton and A. G. Barto, 2018, MIT Press. Copyright 2018, 2020 by Richard S. Sutton and Andrew G. Barto. Reprinted with permission.}
	\label{fig:basic_rl}
\end{figure}

This being a problem of action decisions in a discrete sequence, finite Markov decision processes (MDPs) serve as the mathematical foundation of RL. A MDP describes the probability of ending up in state $s'$ and obtaining reward $r$ when taking action $a$ in state $s$:

\begin{equation}\label{mdp} p(s',r|s,a) \doteq Pr\lbrace S_{t+1}=s',R_{t+1}=r | S_t=s,A_t=a\rbrace.  \end{equation}

This is the central description of the process dynamics of a finite MDP. Based on this information, other quantities can be calculated, such as state-transition probabilities or expected rewards for state-action pairs \cite[p.~63]{Sutton2018}.

The RL agent tasked with maximising the cumulative rewards can be implemented in any of a wide variety of approaches that may or may not include a model of the environment when computing the state transition probabilities. In some cases, the agent may be programmed to learn a control policy optimisation without considering any potential model of the world. In the specific case of model-based reinforcement learning, so far the focus has been into defining predictive models that would allow the agent to guess the best action or sequences of actions to be performed \cite{kaiser2019model}. 

We argue that, in both model-based and model-free approaches, the definition of the RL software library with properly specified software abstractions would add sufficient expressive flexibility to both simplify the formulation of the RL problem, and, when ready, the deployment of the RL model in a real environment.

\subsection{OpenAI Gym Environments}

The OpenAI Gym initiative has created an interface for the interaction of RL agents with RL environments\cite{Brockman16}. This interface is being widely used, and is useful to adopt when creating RL environments for re-use and publication.

A compatible OpenAI Gym enviroment is a Python class inheriting from \texttt{gym.Env} and implementing, as a minimum, the methods \texttt{reset} and \texttt{step}, where the \texttt{reset} method takes no arguments and returns an \texttt{observation} (an OpenAI Gym object representing the current state), and where the \texttt{step} method takes the argument \texttt{action} and returns a tuple of the observation and the reward value after taking the action.

To be usable in a standardised way, an OpenAI Gym environment should employ \texttt{gym}'s package registration mechanism.

\section{Approach}\label{approach}

For a researcher tasked with creating a simulation-based OpenAI-Gym-compatible environment based on a domain model, we propose the following cyclic workflow:

\begin{enumerate}
    \item Develop the domain model in a project which is separate from the simulation and environment specifics.
    \item Determine the research question for which an RL-problem-specific environment should be created.
    \item \label{item:makeenv} For the environment, create a package or project for above research question, separate from the domain model. Using Sim-Env functionality, implement the \texttt{SimulationInterface} and use \texttt{make\_step} to define the environment.
    \item If desired, add orthogonal dimensions to the domain using the extension mechanism \texttt{expose\_to\_plugins}.
    \item Employ the environment or environments to carry out the RL experiment or experiments.
    \item If required, re-iterate from one of these steps, for instance, swap out the environment for a different one, not needing to touch the model (Step \ref{item:makeenv}).
\end{enumerate}

\begin{figure}
    \centering
    \includegraphics[scale=0.35]{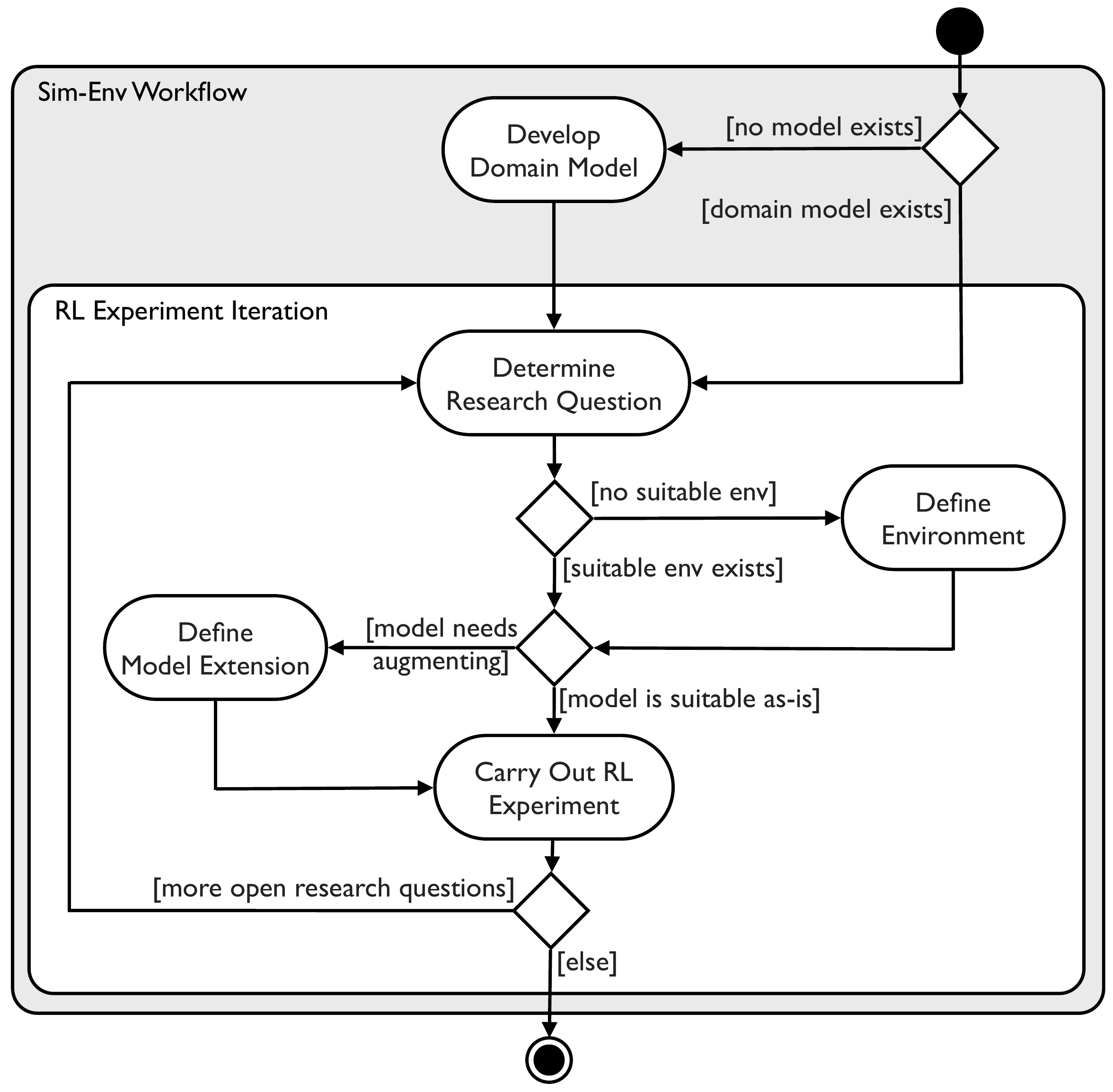}
    \caption{The Sim-Env workflow for iterative simulation-based RL research.}
    \label{fig:workflow}
\end{figure}

Above workflow enables the researcher to create and publish the generic domain model separately from the environment or environments, while still being able to publish any of the resulting environments, for instance as domain-specific baselines. In the latter, one may choose to include the used domain model, or reference it as a dependency.

This workflow minimises the amount of custom environment-specific code, which is particularly useful if the researcher is tasked with a non-trivial matrix of research questions. This independence also allows for clean and maintainable model code. See \ref{showcase} for an object-oriented example model.

With respect to change management, we note that the domain model, as well as each of the environments, should be versioned separately.

\subsection{Guiding Principles}\label{philosophy}

Sim-Env has been developed with the following goals in mind:

\begin{itemize}
\item \textbf{Convenience for simple cases, extensibility for complex cases.} Common RL problems should be very straightforward to implement, but a large number of more complex use cases can also be created in a clean fashion that does not violate the other design considerations.

\item \textbf{Separation between domain model and RL problem environment.} The researcher should be able to model and simulate the domain independently of specific design choices that would have to be made for a particular RL problem environment. This minimises perceived and actual environment-related constraints that may otherwise influence and limit the model's applicability.

\item \textbf{Flexible agency.} There should be no precluded perspective of which modeled entities will act as agents, and which decisions will be handled by a RL algorithm. This enables the researcher to re-use the same model for a greater variety of research questions, such as a combination of rule-based and machine-learning-based principles \cite{chengtrajectory}.

\item \textbf{Extensibility.} It should be possible to update the RL problem without having to change the model's code. For instance, the action and observation spaces can be changed on a finalised model, as can be the reward function. It is also possible to provide any function with plug-in hooks that can be extended without changing the model's code or the environment's code.
\end{itemize}

\subsection{Underlying Technologies}\label{tech}

In order to generate broadly usable environments, we use the interfaces and functionality of the OpenAI Gym library \cite{Brockman16}, such as representations of observation and action spaces, as well as OpenAI's environment registration mechanism.

%the continuation concept should be explained in a background section.
In addition, we use Python's \texttt{threading} package, managing turns in order to create the necessary unification of the perspectives of the domain model and the OpenAI environment user:
\begin{itemize}
	\item \textbf{Domain perspective.} The simulation, which orchestrates the domain model, is in the lead and contains the main loop. It determines when to give entities of the domain model (i.e. potential agents) the opportunity to perform an action. From the domain perspective, the environment user's logic is located inside the body of the action decision function or method.
	\item \textbf{Environment user perspective.} The agent is in the lead and agent-focused code contains the main loop. By choosing an action, the agent gives the environment (i.e. the simulation) the opportunity to react. From the environment user perspective, the domain logic is located inside the body of the \texttt{reset} and \texttt{step} methods.
\end{itemize}

As mandated by the guiding principles, this has been implemented in a manner that neither changes the OpenAI Gym interface nor requires the user to deal with callbacks or similar API complexity. In order to keep interfaces for both sides as simple as calling one function each, the concept of continuations has been employed (in spirit, not as a Python construct), through interlocking threads.

Continuations are a concept in computer science that reifies the control state of a computer program. By turning the way that a program can continue into a first-class object, continuations provide sufficient expressiveness to implement many control patterns like breaks or returns, throwing and catching exceptions, coroutines and generators \cite{moura2009revisiting}.

The decision function jumps to the continuation of whatever comes after the environment's current call to the step function (usually the remainder of the current and beginning of the next iteration of the environment user), while the step function jumps to the continuation of whatever comes after the domain simulation's call to the decision function (also usually the remainder of the current and beginning of the next iteration, only this time of the domain simulation code).

\begin{figure}
    \centering
    \includegraphics[width=\linewidth]{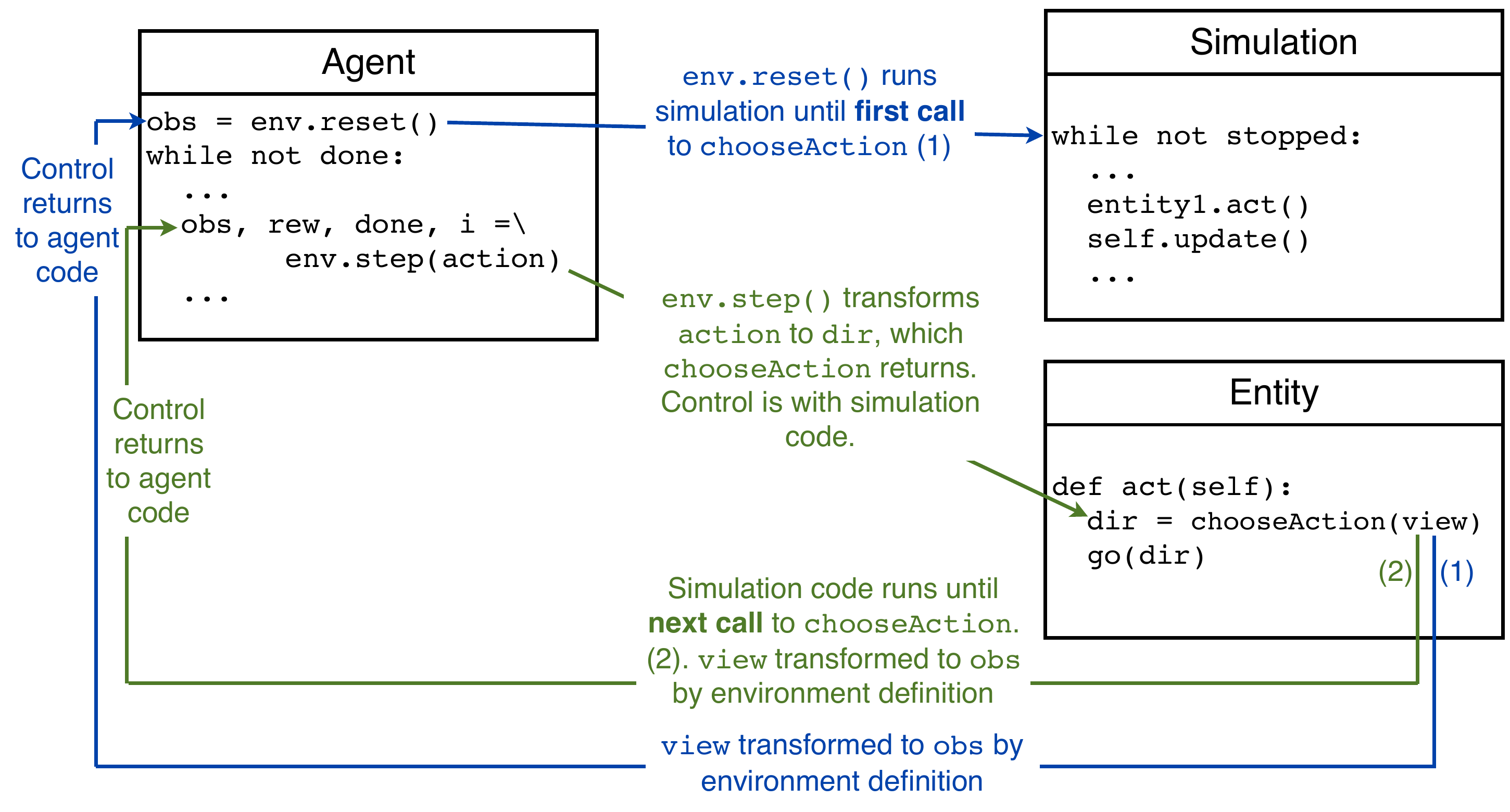}
    \caption{Internals of the unification of environment user and simulation perspectives. From the environment user perspective, all simulation code runs inside \texttt{reset} and \texttt{step}. From the simulation perspective, all of the environment user's code runs inside \texttt{chooseAction}.}
    \label{fig:persp}
\end{figure}

\begin{figure}
    \centering
    \includegraphics[width=\linewidth]{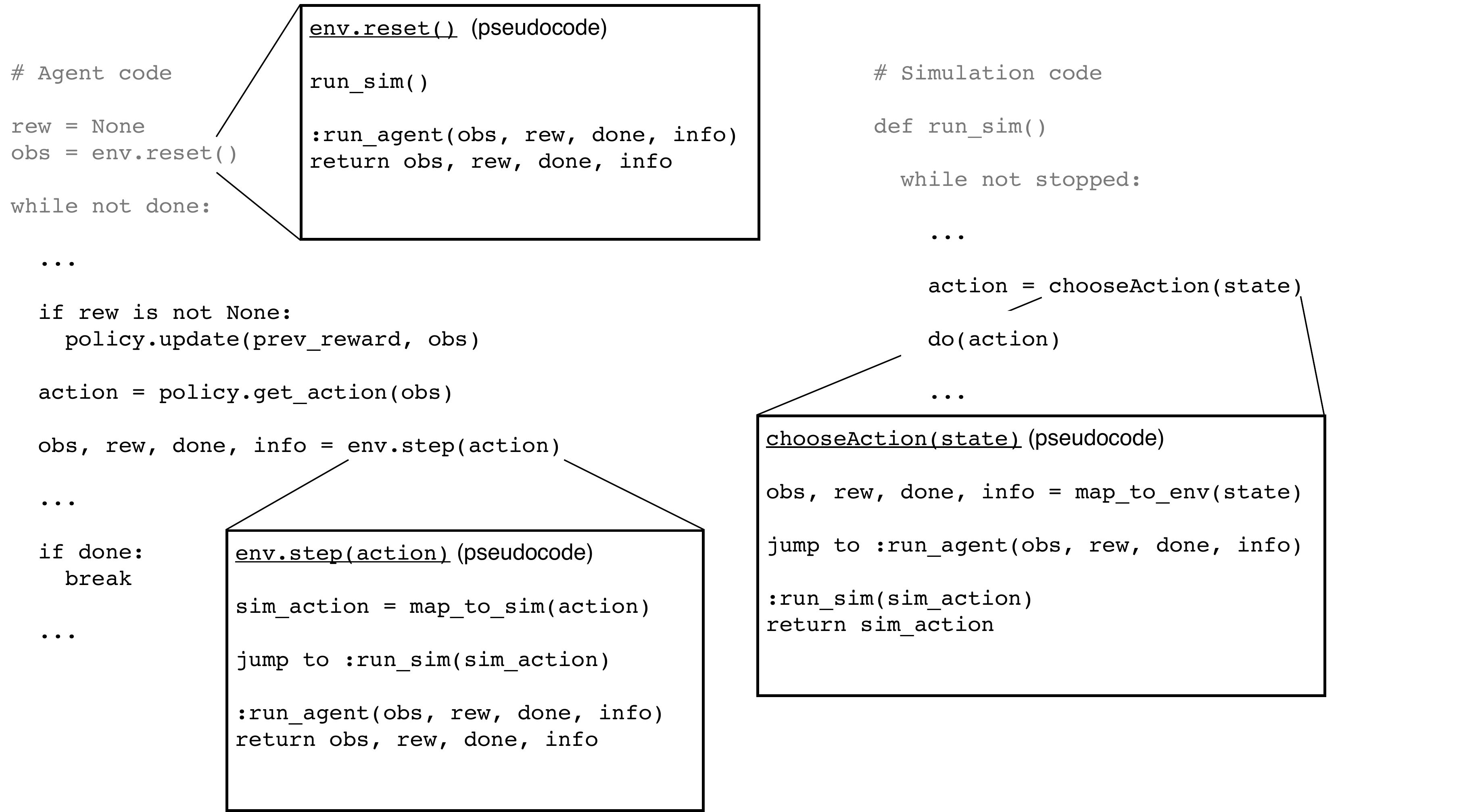}
    \caption{How perspective-switching works. The jump marks starting with colon (\texttt{:}) signify the beginning of the respective continuations.}
    \label{fig:cont}
\end{figure}

% \begin{figure}[htbp!]
%     % \vspace{-0.3cm}

%     \centering
%     \includegraphics[width=0.65\textwidth]{fig/perspectives.pdf}
%         % \vspace{-0.3cm}

%     \caption{
%     The overall architecture with labeled components. 
%     } \label{fig:arch}
%         % \vspace{-0.4cm}

% \end{figure}

\subsection{API Design}\label{api}

% \begin{figure}
%     \centering
%     \includegraphics[scale=0.7]{fig/classes_SimulationInterface.eps}
%     \caption{Abstract base class \texttt{SimulationInterface}}
%     \label{fig:simenvcls}
% \end{figure}

% SIMENV INTRO V1.0 from Research Directions document
The environment-generating interface is located in the \texttt{sim\_env} module. This module contains several functions and class definitions which set up, register and run simulations as an OpenAI Gym environment. As shown in the example below, a compatible simulation needs to implement the \texttt{SimulationInterface}:

\goodbreak
% SIMENV EXCODE V1.0 from Research Directions document
\begin{verbatim}
from gym_fin.envs.sim_interface import SimulationInterface
import my_model

class ExampleSimulation(SimulationInterface):
  def __init__(self):
    self.entity = None
    self.should_stop = False
    ...
  def reset(self):
    """Return simulation to initial state."""
    self.entity = my_model.ActingEntity()
    self.should_stop = False
  def run(self):
    """Run the simulation loop. Terminate at end of an episode."""
    while (not self.should_stop):
        if (entity.should_go_left()):
            entity.go_left()
        else:
            entity.go_right()
        ...
  def stop(self):
    """Tell the simulation to stop/abort the episode."""
    self.should_stop = True
\end{verbatim}

This simulation class is also the right place for any simulation-specific housekeeping. An environment is then defined by using the \texttt{make\_step} decorator on a decision function of the model:

\begin{verbatim}
class ActingEntity:
  [...]
  @make_step(
    observation_space=spaces.Box(len(grid_w * grid_h)),
    observation_space_mapping=lambda self: self.to_obs(),
    action_space=spaces.Discrete(2),
    action_space_mapping={0: True, 1: False},
    reward_mapping=lambda self: 1 if self.active else 0,
  )
  def should_go_left(self):
    # Hard-wired or rule-based agent behaviour which runs
    # if no environment for this decision is created.
    return random.choice([True, False])
\end{verbatim}

The parameters to the \texttt{make\_step} decorator define the generated Gym environment. The decorated decision function or method (here \texttt{should\_go\_left}) is replaced by the RL agent's action-taking. It returns the action passed to \texttt{env.step} to the domain model.

% SIMENV DECORATORS V1.0 from Research Directions document
The decision points marked by the \texttt{make\_step} decorator are registered automatically when importing \texttt{env\_sim}, and an environment for each one these registered decision points is created (in \texttt{sim\_env.generate\_env}), each registered with the OpenAI Gym framework and available through \texttt{gym.make}.

The \texttt{sim\_env} module also provides a simple plugin system, where any function in the simulation code decorated with \texttt{expose\_to\_plugins} can at a later point be extended with additional functionality without having to touch the simulation's code base. One example would be to swap out one financial asset management model for another. This compositionality is orthogonal to the inheritance tree of the simulation classes, making evaluation of different combinations of setups easier.

In order to separate the \texttt{SimulationInterface} subclass from the domain model, we advise to keep it in a separate module from which to import and reference the relevant portions of the domain model.

To separate domain-model-related code from the simulation class or environment definition, these decorators can also be applied from outside the domain model:

\begin{verbatim}
from gym_fin.envs.sim_env import make_step
import my_model

wrap = make_step(
    observation_space=spaces.Box(len(grid_w * grid_h)),
    observation_space_mapping=lambda self: self.to_obs(),
    action_space=spaces.Discrete(2),
    action_space_mapping={0: True, 1: False},
    reward_mapping=lambda self: 1 if self.active else 0,
)
my_model.should_go_left = wrap(my_model.should_go_left)
\end{verbatim}

\subsection{Extensions}\label{extensions}

Many RL research problems require the researcher to deal with a matrix of experiment parameters and/or sweeps across a problem space. While sweeps are well-supported on the side of parametrising RL algorithms, this is less so for the environments that are used to evaluate the RL algorithms.

If a researcher has to vary the complexity of the environment to, for instance, evaluate different RL algorithms with different levels of complexity for a particular problem, there are two choices:

\begin{itemize}
    \item Branch off a new version of the domain model that is the basis of the environment, and apply the necessary changes to it locally.
    \item Make the domain model extensible, so behaviour can be changed without modifying the domain model's code base.
\end{itemize}

The former approach has the disadvantage of creating duplicate code which may become difficult to maintain in the future.

The latter approach is made simple by Sim-Env's \texttt{expose\_to\_plugin} decorator. By applying this decorator to the relevant functions or methods (either from outside or inside the domain model's code), they are registered for plugins or other modifications of their behaviour.

The three possible ways to extend a registered function are by attaching "before", "after" or "instead" handlers. Below conceptual example demonstrates the basics of the extension mechanism.

\begin{verbatim}
import gym_fin.envs.sim_env as se 

def my_func(input):
  """Let's assume my_func is part of a
  black-box domain model in another module"""
  return input + 1

# Non-extended my_func example:
print(my_func(2))  # prints 3

# Making my_func extendable
my_func = se.expose_to_plugins(my_func)

# A simple "before" handler
def before_func(input):
  print(f"got {input}")
se.attach_handler(before_func, my_func, "before")

print(my_func(2))  # prints "got 2" and 3

# A handler can intercept and change input or output
# by returning a ChangedArgs or ChangedResult object.
se.remove_handler(my_func, "before")
def manipulate(input):
  return se.ChangedArgs(input * 2)

print(my_func(2))  # prints 5

# Similarly, handlers can be registered at positions
# "after" and "instead" for functions or methods,
# as well as "before_next" and so on for iterables.
\end{verbatim}

In our experience, this extension technique works well for most well-structured domain models. It allows the researcher to keep the specifics of the experiment design separate from the core domain model code and allows for managing parametrisation using a \textit{design of experiments} package of their choice (for example \texttt{pyDOE2}).
\section{Evaluation}\label{eval}

In this section, we describe our experiences with the proposed workflow and tool, and demonstrate them in a typical use-case.

\subsection{Research Applications}\label{application}

Sim-Env has been used by PhD students and Data Scientists to create simulations to use in RL-related research. In PhD research, it has been used to create and evaluate different approaches of applying RL to business processes in the financial services sector. \cite{FinBaseSim,PensionEnv} Data Scientists have used the package in experimenting with and evaluating the potential viability of approaches of debtor management \cite{MAAT}.

Previously, domain model versions and experiment-specific code had been clumped into the same code base, as the combination of them was regarded as "the RL environment." Although the resulting proliferation of versions for the various research avenues is somewhat manageable through version control techniques such as feature branches, significant effort still was necessary in order to propagate one fix or improvement in the domain model to all relevant environments. We also found ourselves in the dilemma between keeping the models the same while having to implement experiment-specific model behaviour, which led to over-parametrising the model with behaviour that was needed only in one or two experiments, while irrelevant in the rest of the experiments. The Sim-Env workflow and Python package, by cleanly separating the domain model from environment specifics, have been found to help managing the various avenues of research cleanly, and allowing for a significantly higher fraction of shared code.

\subsection{Showcase}\label{showcase}

In this section, we will demonstrate the workflow and package by applying it to a small experimental problem. This particular problem is simple enough to be handled without Sim-Env or with custom parametrisation. But we still hope that it makes clear how the techniques are relevant for larger and more complex experimental problems.

\begin{figure}
    \centering
    \includegraphics[scale=0.55]{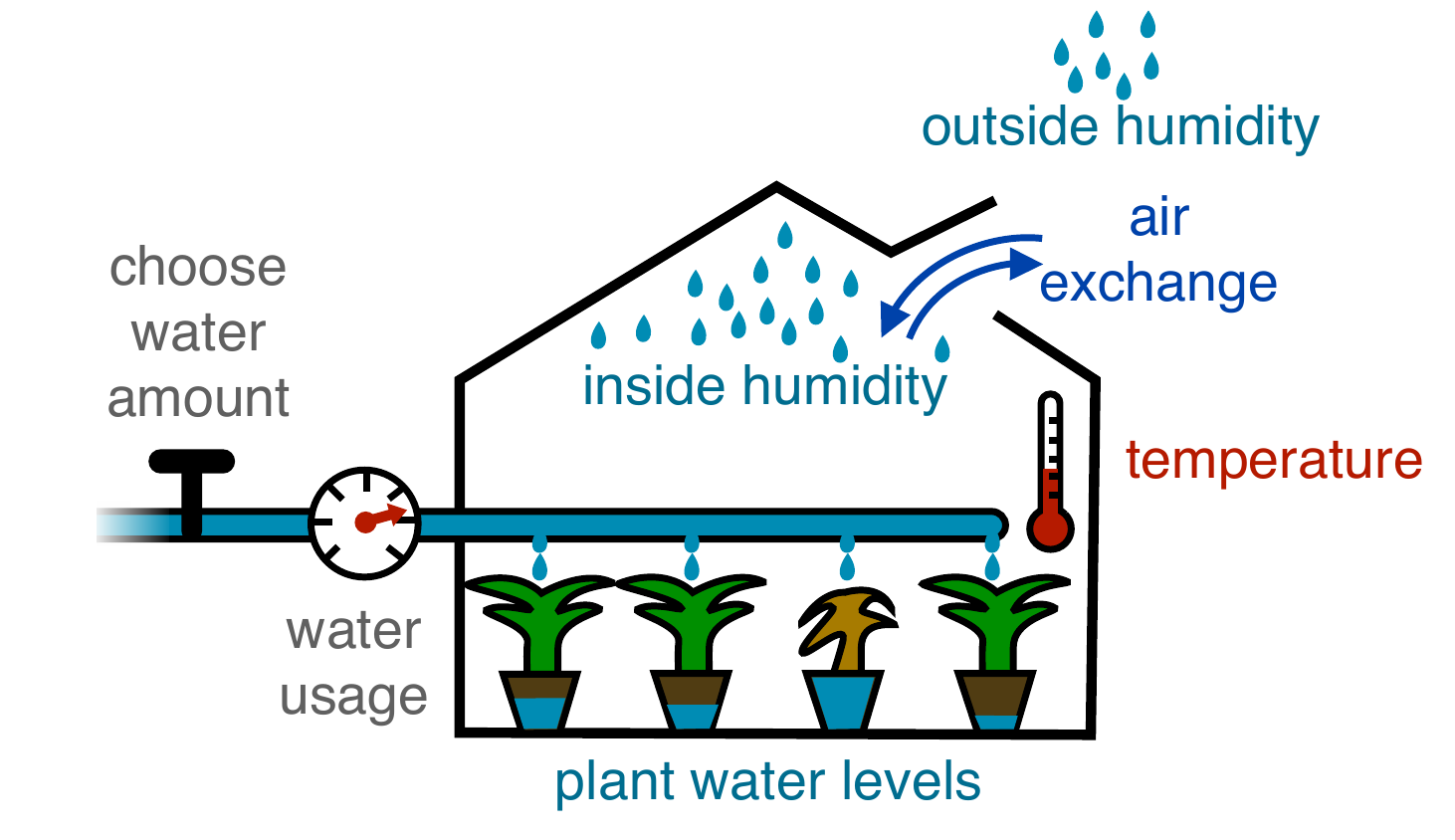}
    \caption{Main components of the Greenhouse example.}
    \label{fig:greenhouse}
\end{figure}

\subsubsection{Step 1: Develop the domain model.}
Our domain model will be a greenhouse watering system. For brevity, we are using drastically simplified physical models:

\begin{itemize}
    \item Each day's temperature is randomly chosen from a uniform distribution between \SI{15}{\celsius} and \SI{35}{\celsius}. % \usepackage{siunitx}
    \item The daily air exchange (and humidity exchange) is fixed at $20\%$, where outside relative humidity is constant at $60\%$.
    \item The daily evaporated mass of water $m_e$ is governed by $m_e = \frac{m_lT}{100 - 100\phi}$, where $m_l$ is the liquid water available in plants, $T$ is the temperature in \si{\celsius}, and $\phi$ is the relative humidity of the greenhouse air.
\end{itemize}

This example's code can be found in supporting material S1.

\begin{figure}
    \centering
    \includegraphics[scale=0.4]{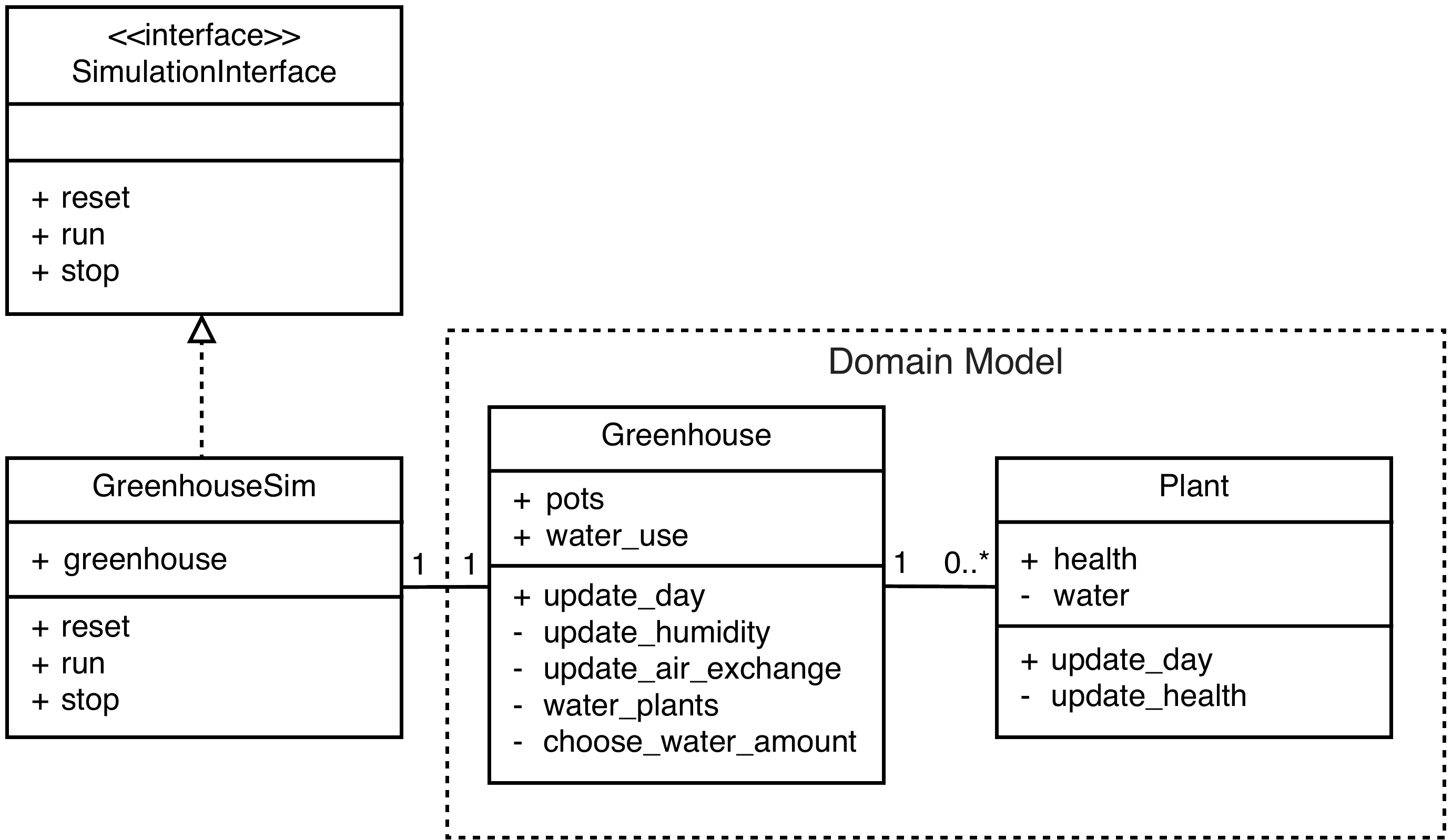}
    \caption{UML representation of the Greenhouse example.}
    \label{fig:greenhouseuml}
\end{figure}

We made this model more interesting by having \texttt{choose\_water\_amount} overwater the plants by default. Running the simulation shows this:

% \goodbreak
\begin{verbatim}
>>> from model import Greenhouse
>>> g = Greenhouse()
>>> for day in range(10):
...   alive = [p.health > 0 for p in g.pots]
...   print(f"day {day} alive: {sum(alive)}, "
...         f"dead: {len(alive)-sum(alive)}")
...   g.update_day()
... 
day 0 alive: 200, dead: 0
day 1 alive: 200, dead: 0
[...]
day 8 alive: 189, dead: 11
day 9 alive: 149, dead: 51
\end{verbatim}

\subsubsection{Step 2: Define the separate environment.}
This module contains the code needed to run the simulation (extending \texttt{SimulationInterface}) and the environment definition.
Please refer to supporting information S2 for this step's source code.

We can now access our environment by importing and instantiating its class \texttt{enf\_def.env\_cls}.

\subsubsection{Step 3: Optionally extend the domain model.}
This is an example for extending an existing domain model for the purposes of a specific experiment.

\begin{verbatim}
# Added to env_def.py before generating env_cls:
from gym_fin.envs.sim_env import expose_to_plugins, attach_handler

def new_air_exchange(self):
  """Faster air replacement with higher temperatures"""
  factor = (self.temp - 15) / 20
  self.humidity += factor * (self.outside_humidity - self.humidity)

Greenhouse.update_air_exchange = \
  expose_to_plugins(Greenhouse.update_air_exchange)
attach_handler(
  new_air_exchange,
  "model.Greenhouse.update_air_exchange",
  "instead"
)
\end{verbatim}

To extrapolate the possibilities at this point, we could also change the function \texttt{new\_air\_exchange} to rely on a newly defined decision function for regulating ventilation. This would be defined as another entry point for a new environment (using \texttt{make\_step}), enabling a cooperative multi-agent setting. The same means would also enable us to combine this new ventilation action with the existing watering action for an extended single-agent setting, creating a new environment that unifies those actions in a two-dimensional action space.

\subsubsection{Step 4: Use the environment.} The generated environment uses the OpenAI Gym interface and can therefore be used as as any other environment, as can be seen in the code fragment below.

\begin{verbatim}
from random import random
import gym

def rand_policy(obs):
    return random()

env = gym.make("env_def:Greenhouse-v0")
obs0 = env.reset()
obs1, rew1, done, info = env.step(rand_policy(obs0))
obs2, rew2, done, info = env.step(rand_policy(obs1))

# day 0 alive: 200, dead: 0
# day 1 alive: 200, dead: 0
# day 2 alive: 182, dead: 18
\end{verbatim}

The example given in this section should make evident how environments with different properties can be defined without needing to adapt the domain model. If the domain model requires corrections, these are easily propagated to all dependent environments, usually without requiring any code changes. If an experiment requires specific adaptations that deviate from the canonical domain model, those are implemented using the extension mechanism shown.

\section{Related Work}\label{related}

The \textit{SuperSuit} Python library provides wrappers for increasing compatibility of existing RL environments \cite{terry2020pettingzoo}. Some features it offers are the transformation of action and observation spaces, as well as environment vectorising and multi-agent interfacing. While it shares some principles with our present work, like not requiring to adapt existing modules, it differs in that its goal is not the creation of new environments as such, but enabling more flexible use of already existing environments, in which SuperSuit and Sim-Env complement each other.

The \textit{MASON} simulation toolkit offers a variety tools for efficient modelling, execution and analysis of agent-based models (among others). While it offers interoparability with a variety of technologies, it still does not tap into the universe of OpenAI-Gym-compatible agents \cite{luke2018mason}.

Zamora et al. present an interface that makes it possible to expose robotics simulations based on ROS and Gazebo as an OpenAI Gym environment \cite{zamora2016extending}. While this toolset combines a modelling framework for robotics problems with the ability to create environments, its focus lies on the robotics domain, and unlike Sim-Env, it is not sensible to use it in any other domain.

\textit{OffWorld Gym} is an API and online service which uses the previously mentioned interface to provide robotics environments both as simulations as well as in the real-world \cite{offworldgym}. While strictly a mere application of Zambora et al.'s work, it showcases the flexibility that can be gleaned by abstracting over RL environments. 

The \textit{MushroomRL} library, similarly to Sim-Env, strives to accelerate RL research by making it possible to combine existing building blocks \cite{deramo2020mushroomrl}. Unlike Sim-Env, however, it focuses on flexibly applying RL algorithms to existing environments. It does not help the issue of creating new RL environments while re-using existing domain models.

As stated in the introduction, there exists extensive work researching complex systems through simulations, % \cite{Haase2009,schreinemachers11}
much of which not does not yet include the aspect of self-learning RL agents. The relation to the present work lies in the possibility to re-use their domain models without the need of major adaptations.
\section{Conclusions and Future Work}\label{future}

In this paper, we have illustrated the Sim-Env workflow and library. Our approach decouples domain model maintenance from RL environment maintenance and thereby aids the acceleration of simulation-based RL research. We expect Sim-Env to contribute as an RL research tool which not only makes it easier but also encourages users to develop environments that use the de-facto standard OpenAI Gym interface and that are more broadly re-usable (as are their components).

The present work's limitations include:
\begin{enumerate*}
    \item\label{limit1} a lack of direct support of multi-agent reinforcement learning (MARL) and    \item\label{limit2} a lack of direct support of alternative interfaces such as vectorised environments, and
    \item\label{limit3} the use of interlocking threads instead of Python-native continuations.
\end{enumerate*}

To address \ref{limit1}. and \ref{limit2}. we plan to extend Sim-Env API compatibility towards PettingZoo \cite{terry2020pettingzoo} and goal-based environments (\texttt{gym.GoalEnv}) \cite{Brockman16}. We intend to mitigate \ref{limit3}. with native coroutines or, if achievable, native continuations.

In order to further lower the barriers of entry to simulation-based RL, specifically in the financial services sector, we intend to create a domain base model called \textit{Fin-Base} which contains many building blocks of creating a simulation of interactions between financial entities, that is to be exposed as RL environment using Sim-Env.

% Ack needs to be run-in heading in LNCS:
\subsubsection{\ackname}
This work is part of the first author's PhD research and is supported by APG Algemene Pensioen Groep N.V. The Sim-Env Python package relies on a minimum of dependencies. It is available under the Apache 2.0 open source license at \url{https://github.com/schuderer/bprl}.

\bibliographystyle{splncs04}
\bibliography{bibliography,references}
% \bibliography{references.bib}

\end{document}

\begin{center}
  \Large\bfseries\boldmath
  Supplementary Material:\\
  Sim-Env: Decoupling OpenAI Gym Environments from Simulation Models Title
\end{center}
%

% LNCS: If a paper includes an Appendix, it should be placed in front of the references. If it has been placed elsewhere, it will be moved by our typesetters. If there is only one, it is designated “Appendix”; if there are more than one, they are designated “Appendix 1”, “Appendix 2”, etc.
% Appendixes should be referred to in the text. The content of an appendix is con- tained within the sections subordinated to the major heading “Appendix”. The lan- guage and styling rules for the text also apply to the appendixes. The form of number- ing of tables, figures, and equations in an appendix should be the same as in the body of the article, continuing the numbering used there.
% \section*{Appendix}
% \appendix
% \renewcommand{\thesection}{Appendix \arabic{section}}
\renewcommand{\thesection}{S\arabic{section}}
\renewcommand{\thefigure}{S\arabic{figure}}

% \section{}\label{app:simul_example_code}

\section{Domain Model Code of Showcase Example}

This is the source code of the greenhouse watering system model.

\begin{verbatim}
# model.py
from math import exp
from random import random, seed

class Greenhouse:
  def __init__(self, seed_val=None):
    if seed_val is not None:
      seed(seed_val)
    self.pots = [Plant(self) for _ in range(200)]
    self.water_use = 0
    self.temp = 20 # degrees C
    self.humidity = 0.6  # relative %
    self.outside_humidity = 0.6
    self.size = 2400  # cubic metres
    
  def update_humidity(self):
    evaporation_factor = self.temp / 100 * (1 - self.humidity)
    evaporated = 0
    for plant in self.pots:
      evaporated += evaporation_factor * plant.water
      plant.water = max(
        0,
        plant.water - evaporation_factor * plant.water
      )
    max_saturation = 0.20  # kg/m3 at 22 degrees C
    saturation = \
      self.humidity * max_saturation + evaporated / self.size
    self.humidity = saturation / max_saturation
    
  def update_air_exchange(self):
    # 20% air exchange per day
    self.humidity = 0.8 * self.humidity + 0.2 * self.outside_humidity

  def choose_water_amount(self):
    return 200

  def water_plants(self):
    water_amount = self.choose_water_amount()
    for plant in self.pots:
      plant.water += water_amount / len(self.pots)
    self.water_use += water_amount

  def update_day(self):
    self.temp = min(35, max(15, self.temp + int(random()*5) - 2))
    self.update_humidity()
    self.update_air_exchange()
    for plant in self.pots:
      plant.update_day()
    self.water_plants()

class Plant:
  def __init__(self, greenhouse):
    self.greenhouse = greenhouse
    self.water = 2  # litres in pot
    self.health = random() * 0.8 + 0.1
    self.req_water = 0.3 * random() + 0.1

  def update_health(self):
    if self.health == 0:
      return
    if self.water <= 0 or self.water > 3:
      self.health = max(0, self.health - 0.25)
    else:
      self.health = min(1, self.health + 0.1)

  def update_day(self):
    self.water = max(0, self.water - self.req_water)
    self.update_health()
\end{verbatim}

\section{Simulation and Registration Code of Showcase Example}

This is the code needed to run the simulation (extending \texttt{SimulationInterface}) and of the environment definition. It results in a widely usable OpenAI Gym environment.

\begin{verbatim}
# env_def.py
from gym import spaces
from gym_fin.envs.sim_env import make_step
from gym_fin.envs.sim_interface import SimulationInterface
import numpy as np
from model import Greenhouse

# Create the simulation runner class
class GreenhouseSim(SimulationInterface):
  def __init__(self, seed=None):
    self.seed = seed
    self.greenhouse = None
    self.should_stop = False
    self.day = 0

  def reset(self):
    self.greenhouse = Greenhouse(self.seed)
    self.should_stop = False
    self.day = 0

  def run(self):
    while (not self.should_stop):
      alive = [p.health > 0 for p in self.greenhouse.pots]
      
      print(f"day {self.day} alive: {sum(alive)}, "
            f"dead: {len(alive)-sum(alive)}")
      self.greenhouse.update_day()
      self.day += 1
      if sum(alive) == 0:
        self.should_stop = True

  def stop(self):
    self.should_stop = True


# Define the environment:
def obs_from_greenhouse(g):
    # all spaces normalised to make broadly usable
    t = (g.temp - 15) / 20
    h = g.humidity
    n = min(len(g.pots), 1000) / 1000
    alive = [p.health > 0 for p in g.pots]
    a = sum(alive)/len(alive)
    return np.array([t, h, n, a])

last_water_use = 0
def reward_from_greenhouse(g):
    alive = [p.health > 0 for p in g.pots]
    alive_percent = sum(alive)/len(alive)
    global last_water_use
    water_diff = g.water_use - last_water_use
    water_cost = water_diff / 1000
    last_water_use = g.water_use
    return alive_percent - water_cost

env_def = make_step(
    # all spaces normalised to make broadly usable
    observation_space=spaces.Box(
      low=np.array([0.0, 0.0, 0.0, 0.0]),
      high=np.array([1.0, 1.0, 1.0, 1.0])
    ),
    observation_space_mapping=obs_from_greenhouse,
    action_space=spaces.Box(low=0.0, high=1.0, shape=(1,)),
    action_space_mapping=lambda w: min(max(0, w*1000), 1000),
    reward_mapping=reward_from_greenhouse
)
Greenhouse.choose_water_amount = \
  env_def(Greenhouse.choose_water_amount)

# Create the env class
from gym.envs.registration import register
from gym_fin.envs.sim_env import generate_env
env_cls = generate_env(
  GreenhouseSim(),
  "model.Greenhouse.choose_water_amount"
)
register(
  id="Greenhouse-v0",
  entry_point="env_def:env_cls",
  kwargs={},
)
\end{verbatim}